\documentclass[lettersize,journal]{IEEEtran}
\usepackage{amsmath,amsfonts}
\usepackage{algorithmic}
\usepackage{algorithm}
\usepackage{array}
\usepackage[caption=false,font=normalsize,labelfont=sf,textfont=sf]{subfig}
\usepackage{textcomp}
\usepackage{stfloats}
\usepackage{url}
\usepackage{verbatim}
\usepackage{graphicx}
\usepackage{cite}
\usepackage{multirow}
\usepackage{color}
\definecolor{citeco}{RGB}{0,174,239}
\usepackage[colorlinks,
linkcolor=red,       
anchorcolor=blue,  
citecolor= citeco,        
]{hyperref}
\def\eg{\emph{e.g.\ }}  
\def\ie{\emph{i.e.\ }}

\def\etal{\emph{et al\ }} 

\begin{document}

\title{Modality Prompts for Arbitrary Modality Salient Object Detection}

\author{Nianchang Huang, Yang Yang, ~Qiang Zhang*,~Jungong Han, Jin Huang
	\thanks{Nianchang Huang, Yang Yang and Qiang Zhang are with the State Key Laboratory of Electromechanical Integrated Manufacturing of High-Performance Electronic Equipments, and the Center for Complex Systems, School of Mechano-Electronic Engineering, Xidian University, Xi’an, Shaanxi 710071, China. Email: nchuang@stu.xidian.edu.cn, yang@stud.xidian.edu.cn, and qzhang@xidian.edu.cn. }
	\thanks{Jungong Han is with the Department of Computer Science, University of Sheffield, U.K. Email: jungonghan77@gmail.com }
	\thanks{Jin Huang is with the State Key Laboratory of Electromechanical Integrated Manufacturing of High-Performance Electronic Equipment, Xi’an, Shaanxi 710071, China. Email: jhuang@mail.xidian.edu.cn.}
	\thanks{*Corresponding authors: Qiang Zhang.} }	

\markboth{Journal of \LaTeX\ Class Files,~Vol.~14, No.~8, August~2021}%
{Shell \MakeLowercase{\textit{et al.}}: A Sample Article Using IEEEtran.cls for IEEE Journals}


\maketitle

\begin{abstract}
	This paper delves into the task of arbitrary modality salient object detection (AM SOD), aiming to detect salient objects from arbitrary modalities, \eg RGB images, RGB-D images, and RGB-D-T images. A novel modality-adaptive Transformer (MAT) will be proposed to investigate two fundamental challenges of AM SOD, \ie more diverse modality discrepancies caused by varying modality types that need to be processed, and dynamic fusion design caused by an uncertain number of modalities present in the inputs of multimodal fusion strategy. Specifically, inspired by prompt learning's ability of aligning the distributions of pre-trained models to the characteristic of downstream tasks by learning some prompts, MAT will first present a modality-adaptive feature extractor (MAFE)  to tackle the diverse modality discrepancies by introducing a modality prompt for each modality. In the training stage, a new modality translation contractive (MTC) loss will be further designed to assist MAFE in learning those modality-distinguishable modality prompts. Accordingly, in the testing stage, MAFE can employ those learned modality prompts to adaptively adjust its feature space according to the characteristics of the input modalities, thus being able to extract discriminative unimodal features. Then, MAFE will present a channel-wise and spatial-wise fusion hybrid (CSFH) strategy to meet the demand for dynamic fusion. For that, CSFH dedicates a channel-wise dynamic fusion module (CDFM) and a novel spatial-wise dynamic fusion module (SDFM) to fuse the unimodal features from varying numbers of modalities and meanwhile effectively capture cross-modal complementary semantic and detail information, respectively. Moreover, CSFH will carefully align CDFM and SDFM to different levels of unimodal features based on their characteristics for more effective complementary information exploitation. 
	Experimental results show that by virtue of MAFE, MTC loss and  CSFH, our proposed MAT achieves significant increasements over existing models on those benchmark datasets. 
\end{abstract}


\section{Introduction}\label{sec::Intro}

\IEEEPARstart{S}{alient} object detection (SOD) \cite{10204274} strives to identify the most visually appealing objects within input images, which has been extensively employed in numerous computer vision applications, encompassing tasks like tracking, segmentation, camouflaged object detection \cite{9578707}, and so on.

\begin{figure}[!t]
	\centering
	\includegraphics[width=\linewidth]{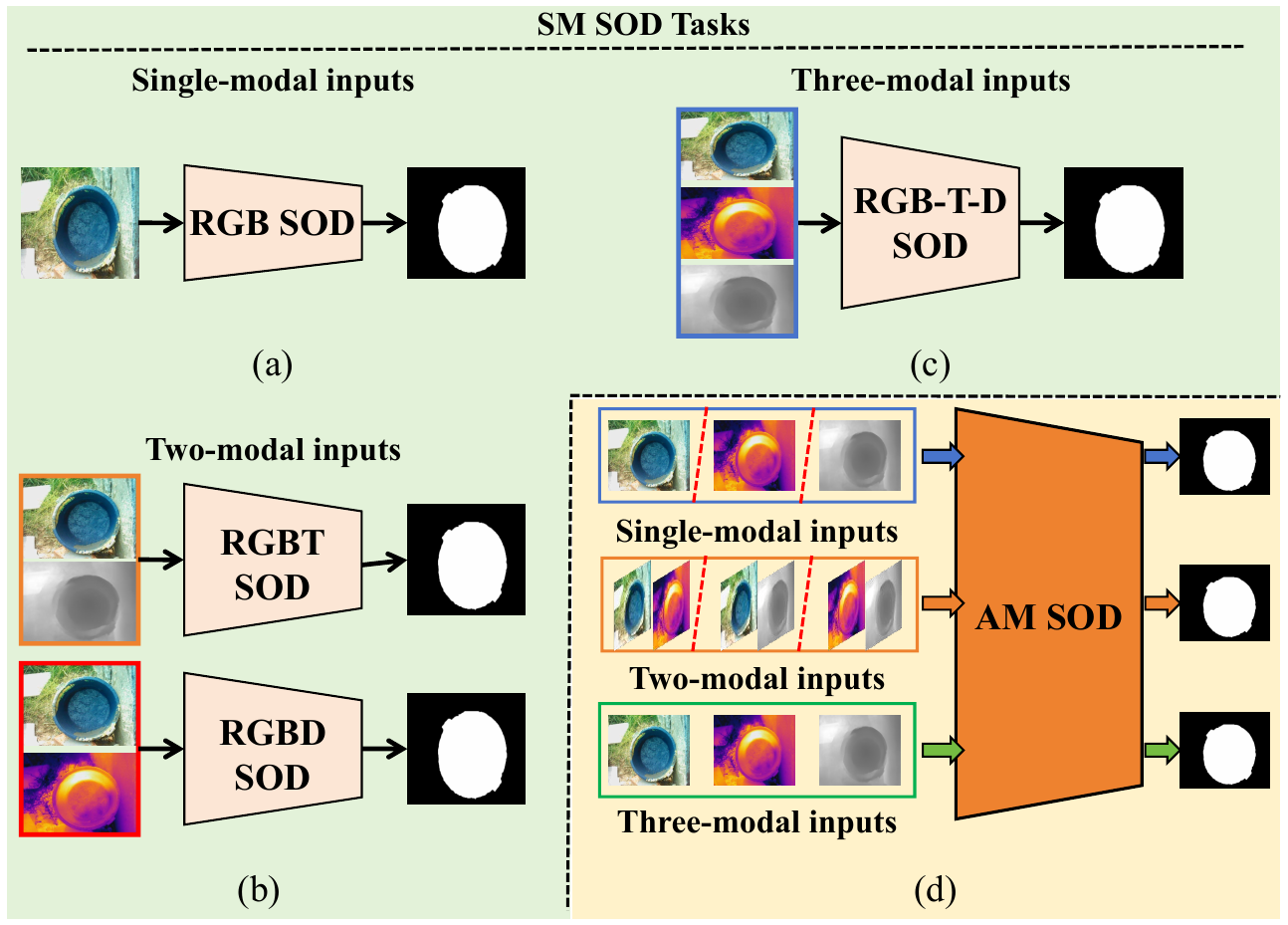}
	\caption{Comparisons of different SOD tasks. (a) Single-modal RGB SOD. (b) Two-modal RGB-D/RGB-D SOD. (c) Three-modal RGB-D-T SOD. (D) A SOD.  }
	\label{fig_AMSOD}
\end{figure}

Early, researchers mainly focus on detecting salient objects from single-modal RGB images (\ie RGB SOD in Fig. \ref{fig_AMSOD}(a)) and have achieved relatively satisfactory results in those simple scenarios \cite{z11,9791375,9745960}. However, visible cameras have many limitations, \eg it cannot capture enough information in low light conditions and loses abundant 3D spatial information. As a result, RGB SOD usually obtains terrible results in those complex scenes. Recently, multimodal SOD \cite{n50, z20, 9931143,RGBTSOD1} has received great attention, which aims to exploit the complementary information among different modalities to break the bottleneck of RGB SOD (\ie Fig. \ref{fig_AMSOD}(b) and (c)). For example, jointly utilizing RGB images and thermal (T) images (RGB-T SOD) to detect salient objects from dark scenes or utilizing RGB images and depth (D) images (RGB-D SOD) for the scenes that targets share similar shapes and colors with the backgrounds. 

Although growing fast, most existing SOD models, including unimodal SOD and multimodal SOD, are designed for some fixed modality types with specific modality numbers (\ie FM SOD). 
If changing their input modalities, such as using RGB images for RGB-D SOD models or D-T images for RGB-T SOD models,  the effectiveness of those FM SOD models will be undermined.
Therefore, they are not suited for applications with the demands of changing modality types, \eg using RGB images for simple scenes,  RGB-T images at night, or RGB-T-D images for dark and complex scenes. For building a more universal SOD model, a novel Arbitrary Modality SOD (AM SOD) has been proposed recently, aiming at detecting salient objects from arbitrary modalities.


Compared with FM SOD, AM SOD has two basic challenges. The first one is more diverse discrepancies among multiple modalities. Specifically, AM SOD usually has to handle more modality types than existing FM SOD models. while each modality has its own characteristics due to different imaging mechanisms, thus leading to more diverse modality discrepancies. 
This poses a significant challenge in effectively extracting discriminative unimodal features over discrepancies across multiple modalities with a limited amount of parameters. Especially, the existing single-stream structures are inadequate in addressing diverse modality discrepancies, while multi-stream structures inevitably lead to a substantial increase in vast parameters as the number of modalities multiplies \cite{k02}.
The second challenge lies in the uncertain number of modalities for the inputs of multimodal feature fusion strategies. Specifically, AM SOD models can receive inputs ranging from one image of RGB/D/T data to two images of RGB-D/RGB-T/D-T data, and even three images of RGB-D-T data, and so on. Consequently, unlike existing FM SOD models that only need to fuse a fixed number of unimodal features, the fusion strategies of AM SOD models must possess the capability to dynamically fuse unimodal features of varying numbers.

Our prior work proposed a preliminary solution, namely the modality switch network (MSN), for AM SOD. In the feature extraction stage, MSN designed a modality switch feature extractor (MSFE) to tackle the diverse modality discrepancies by introducing a modality indicator for each modality. Specifically, MSFE leverages each modality indicator to generate a specific set of modality switch weights for extracting discriminative unimodal features from the images of a certain modality. This enables MSN to utilize a single feature extractor for extracting discriminative unimodal features without incurring any additional parameters. In the fusion stage, MSN presented a dynamic fusion module (DFM) to adaptively fuse unimodal features from varying modality numbers. Specifically, it dedicates a novel Transformer structure that treats the unimodal features of each modality as a token of the cross-attention module and establishes the channel-wise relations of different modalities, achieving dynamic fusion for a varying number of unimodal features.

\begin{figure}[!t]
	\centering
	\includegraphics[width=\linewidth]{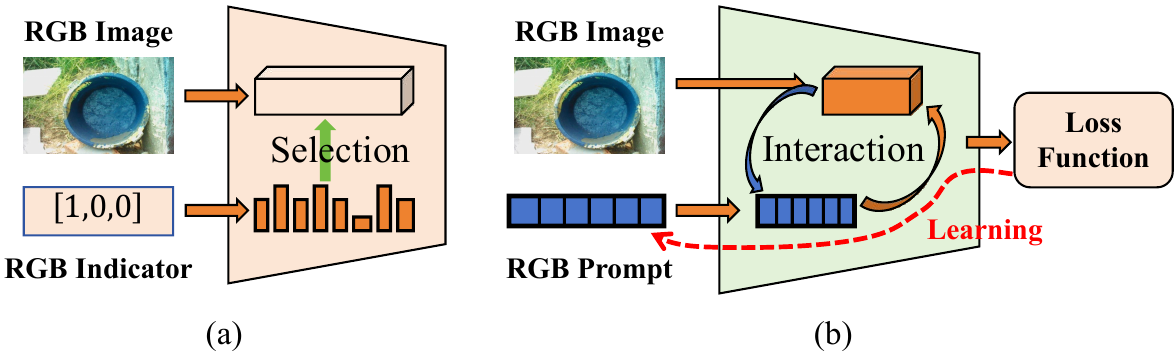}
	\caption{Limitions of MSN. (a) Modality indicators for feature extraction. (b) Modality prompts for feature extraction.  }
	\label{fig_nc}
\end{figure}

Despite its noteworthy advancements, MSN still has some limitations. First, as shown in Fig. \ref{fig_nc}(a), the modality indicators in MSFE are fixed vectors for different modalities (\eg [1,0,0] for RGB modality). Accordingly, the modality switch weights can only select a particular subset of features for each modality. This implicitly narrows the representation space of each modality, as it fails to utilize all available features to comprehensively express the inputs, thus leading to insignificant feature extraction. Secondly, the DFM mainly focuses on establishing the channel-wise relations among different modalities in a dynamic fusion way but ignores their spatial-wise relations. However, the spatial-wise complementarity among different modalities also plays an important role in locating salient objects and recovering their boundaries, thus leading to insignificant cross-modal complementary information exploitation.

This paper will propose a novel modality-adaptive Transformer (MAT) to address the above issues. Recently, prompt learning has emerged as a hot topic in computer vision.  One of the primary functions of prompt learning lies in its ability to fine-tune the feature space learned in the pre-trained data and align them with the characteristics of downstream tasks by learning some prompts. Therefore, no more need to change the parameters of pre-trained models and just learn different prompts with a few parameters, prompt learning can obtain remarkable results for different downstream tasks. As shown in Fig. \ref{fig_nc}(b), this inspires us to specially learn a modality prompt for each modality in the feature extraction stage and exploit these prompts to tune our model's feature space according to the characteristics of the input modalities. Doing so will enable our proposed model to extract discriminative unimodal features from arbitrary modalities via all available features of a single feature extractor, without losing representational ability and increasing the vast amount of parameters. Secondly, in the fusion stage, MAT will explore the channel-wise and spatial-wise relations among the unimodal features from an arbitrary number of modalities in a dynamic fusion way.

Specifically, in the feature extraction stage, a novel modality-adaptive feature extractor (MAFE) is designed to  tackle the diverse modality discrepancies and capture discriminative unimodal features according to the characteristics of input modalities with the aid of corresponding modality prompts. Based on the Transformer structure, MAFE first takes an image of arbitrary modalities with its corresponding modality prompt as the inputs and utilizes the Transformer's self/cross-attention mechanism to establish interactions among unimodal features and modality prompt. Such interactions enable the modality prompt to tune/adjust the characteristics/distributions of unimodal features. Furthermore, a new modality translation contractive (MTC) loss is designed in the training stage to learn more modality-distinguishable prompts. It will pull the features with the same modality prompts toward each other and push the features with the different modalities away from each other by changing inputs' modality prompts. Doing so will let modality prompts effectively adjust the distributions of unimodal features according to the characteristics of input modalities, thus enhancing the discrimination of those unimodal features. 
Eventually, in the testing stage, MAFE can effectively extract discriminative unimodal features with those learned modality prompts according to the characteristics of input modalities just by using one network in a large feature space.

In the fusion stage, a channel-wise and spatial-wise fusion hybrid (CSFH) strategy is further designed in our proposed MAT to dynamically and effectively exploit the complementary semantic and detail information across an arbitrary number of modalities. Specifically, on top of the channel-wise dynamic fusion module (CDFM) proposed in our previous work, a novel spatial-wise dynamic fusion module (SDFM) is further developed. SDFM first transfers the unimodal features of one modality as a token, thus constructing a sequence whose length is the same as the number of modalities. Then, SDFM will exploit the spatial complementarity among input modalities by using the self-/cross-attention mechanism. Finally, given CDFM and SDFM, our proposed CSFH will carefully align them to different levels of unimodal features, since the features from different levels obtain different characteristics \cite{SAA, FPT,Zhang_2022_CVPR}. For low-level features, CSFH employs SDFM for capturing more complementary detail information, while for high-level features, CSFH mainly utilizes CDFM for exploiting their complementary semantic information. Eventually, CSFH captures more complementary information and significantly boosts performance. 

In summary, the main contributions of this work are as follows: 

(1)  We take the initiative to investigate exploiting modality prompts to tune AM SOD model's feature space for extracting discriminative unimodal features from different modalities and simultaneously establishing spatial-wise and channel-wise relations across modalities to effectively capture cross-modal complementary information.  Eventually, our proposed modality-adaptive Transformer can well handle the more diverse modality discrepancies issue and dynamic fusion issue of AM SOD task, thus significantly boosting performance.

(2) We present a novel modality-adaptive feature extractor (MAFE) and a new modality translation contractive (MTC) loss. In the training stage, they can effectively learn modality-distinguishable prompts by generating different image modalities with the aid of different modality prompts. While, in the testing stage, they are able to effectively extract discriminative unimodal features according to the characteristics of different modalities by using those learned modality prompts.

(3) We propose a new channel-wise and spatial-wise fusion hybrid (CSFH) strategy, which can effectively capture complementary detail and semantic information across multiple modalities by using spatial-wise dynamic fusion module (SDFM) and channel-wise dynamic fusion module (CDFM), respectively. Moreover, CSFH exploits more complementary information across modalities by aligning SDFM and CDFM for different levels of unimodal features, thus further improving performance.

\section{Related work} \label{sec::RW}

\subsection{FM SOD}

FM SOD refers to the models that focus on detecting salient objects from the inputs with fixed modalities and modality numbers. Generally speaking, FM SOD tasks mainly consist of single-modality RGB SOD, two-modality RGB-D SOD and RGB-T SOD, three-modality RGB-D-T SOD. Among that, RGB SOD is studied earliest and most completely, especially, deep learning based models significantly boost their performance by a large margin\cite{GPONet,RGB1,9224166,9514538}. Generally speaking, existing RGB SOD models mainly focus on how to exploit context information within the inputs for locating salient objects and recovering their boundaries. For example, Zhao \etal \cite{9957106} proposed a part-whole hierarchies and contrast cues-based network (PWHCNet). PWHCNet first explores the part-whole relational cues across the salient objects and their backgrounds for effectively capturing the context information around salient objects by using Capsule Network (CapsNet). Then, it exploits the contrast cues across foreground and background capsules for complementing part-whole relational cues. By doing so, PWHCNet obtains large improvements in several public datasets.

Recently, multi-modal SOD tasks, including two-modality SOD \cite{ n51, JIN2024123222, z28} and three-modality SOD, have undergone significant advancements since leveraging complementary information across different modalities can break the bottleneck of single-modality RGB SOD in complex scenes, thereby enhancing its practicality in real-life applications. Multi-modal SOD tasks mainly focus on how to effectively exploit cross-modal complementary information across modalities by dedicating different fusion strategies for obtaining more scene information, and how to capture more context information from those fused information to accurately locate and recover salient objects. For example, on top of widely-used linear fusion strategies, \eg element-wise addition and concatenation, Huang \etal \cite{n51} proposed to further investigate the high-order no-linear fusion strategies for establishing no-linear relations among modalities and effectively capturing more complementary information. For that, they proposed a Multi-modal Feature Interaction (MFI) module which explores the linear and no-linear relations across modalities by using weighted-addition fusion and bilinear fusion, respectively. Eventually, their model's performance achieves significant improvements in RGB-D SOD.

Nowadays, FM SOD has made extraordinary progress and existing models can satisfy the diverse needs of some real-life applications. However, their shortcomings are also evident, \ie existing FM SOD models are specially designed or trained for the particular inputs with fixed modality types and modality numbers. If feeding them with other modalities, those models will lose effectiveness. However, based on the concerns of efficiency and energy conservation numerous real-life applications necessitate the ability to change the modalities of their inputs under various conditions.

\subsection{AM SOD}

Before Huang \etal proposes the new task of AM SOD, many researchers have realized the above drawbacks of FM SOD and have made their efforts to address those issues. First, some works try to build an unfiled two-modality SOD model \cite{n30, z27,Un1, HRTransNet, Un2} that can simultaneously detect salient objects from RGB-D images and RGB-T images.  For example, Chen \etal \cite{Un2} presented a modality-induced transfer-fusion network (MITF-Net) for RGB-D and RGB-T SOD. MITF-Net mainly focuses on how to fully explore the complementarity in multi-modality data. For that, it first bridges the semantic gap between single and multi-modality images by designing a modality transfer fusion (MTF) module. Then, it employs a cycle-separated attention (CSA) module to recurrently exploit the complementary information within multi-level features. Finally, MITF-Net optimizes saliency maps' boundaries in the saliency prediction stage. This model obtains good performance on 13 RGB-D and RGB-T SOD datasets. However, those unfiled two-modality SOD models are also limited by the fact that they need to be trained separately on those RGB-D and RGB-T datasets and cannot detect salient objects from RGB-D images and RGB-T images by using one model with the same parameters.

Recently, Jia \etal \cite{jia2023one} proposed an all-in-one SOD model, namely AiOSOD. This model can detect salient objects from three types of data (RGB, RGB-D, and RGB-T) by using one model with the same weight parameters. It first merges depth images or thermal images with RGB images in a batch in an orderly manner by considering them as a special kind of RGB image. This actually unifies single-/dual-image inputs into dual-image inputs, \ie mainly transferring RGB inputs into RGB-RGB inputs. Then, it extracts unimodal features from different modalities via the same weights but different norm layers for detecting salient objects. Although AiOSOD achieves good performances in RGB, RGB-D, and RGB-T datasets. it is specially designed for such three tasks and cannot detect salient objects from arbitrary modalities, \eg Depth/Thermal SOD and RGB-D-T SOD. 

Eventually, Huang \etal proposes the task of AM SOD, which aims at detecting salient objects from the inputs with arbitrary modalities. Specifically, Huang \etal builds a new AM SOD dataset, AM-XD dataset, for training and testing different models. Furthermore, they also propose a preliminary solution, modality switch network (MSN), for such a task. MSN first designs a modality switch feature extractor for extracting unimodal features from arbitrary modalities by generating different weights from some dedicated  modality indicators for different modalities.  Then, MSN presents a dynamic fusion module (DFM) for fusing unimodal features of varying numbers. Finally, MSN achieves good performance on the AM-XD dataset. 

It can be seen that the exploration into AM SOD is merely at its inception.  There are many problems that need to be solved in the AM SOD task. This paper makes further attempts to address its more diverse modality discrepancies and dynamic fusion challenge by proposing a novel modality-adaptive Transformer.

\section{Proposed Model}\label{sec::Model}

\begin{figure*}[!t]
	\centering
	\includegraphics[width=\linewidth]{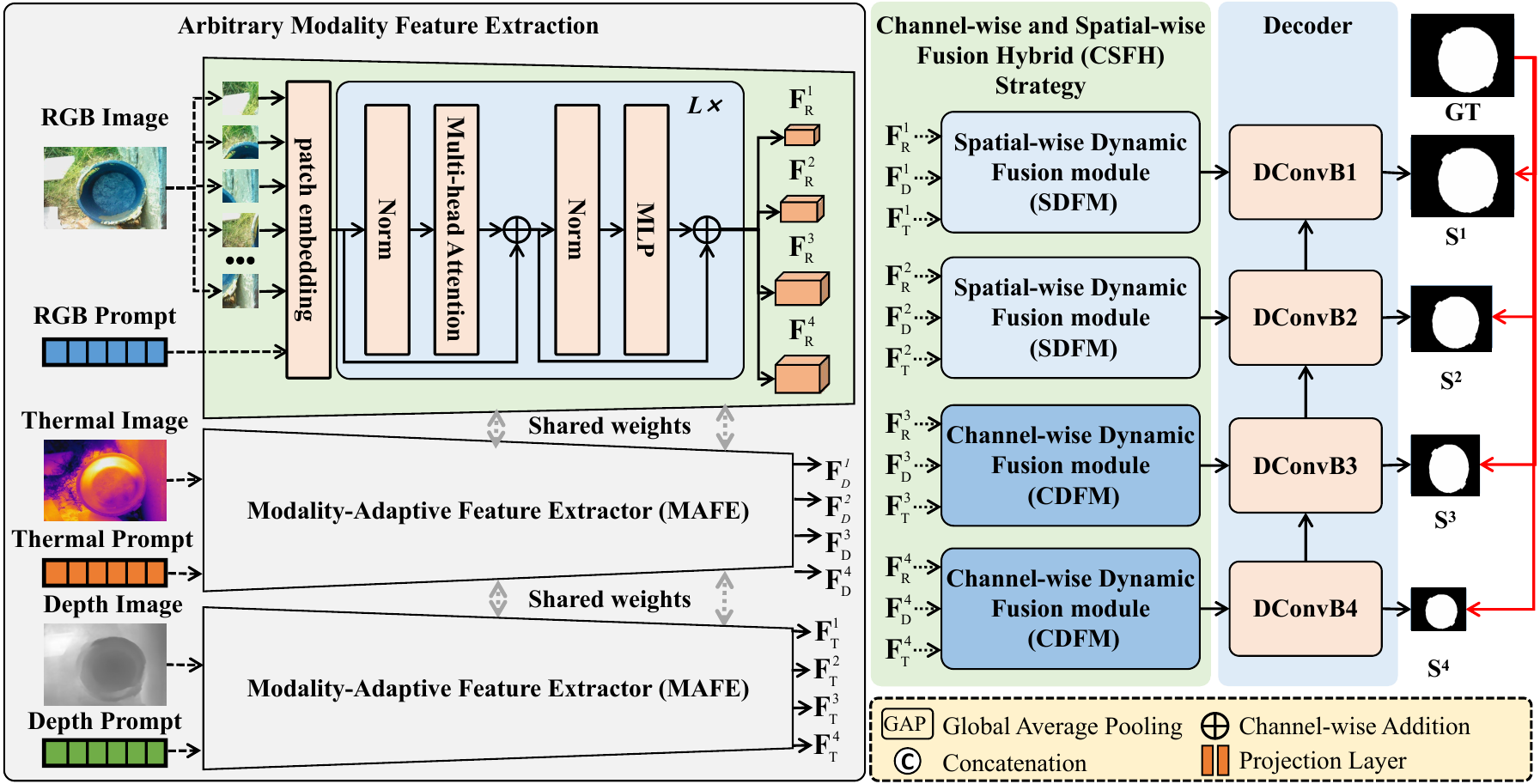}
	\caption{Framework of our proposed modality-adaptive Transformer (MAT). We employ the three-modality RGB-D-T inputs as illustrative examples. First, the modality-adaptive feature extractor  (MAFE) will receive an arbitrary modality image along with its corresponding modality prompt as inputs and then proceed to extract four distinct levels of unimodal features. After obtaining the unimodal features of RGB modality, depth modality, and Thermal modality, respectively, the channel-wise and spatial-wise fusion hybrid (CSFH) strategy will fuse these unimodal features by aligning SFDM and CFDM for different levels of unimodal features. Finally, MAT will leverage a saliency decoder to predict salient objects based on these fused features.   }	
	\label{fig_Frame}
\end{figure*}

\subsection{Overall Framework}

The overall framework of our proposed modality-adaptive Transformer (MAT) is shown in Fig. \ref{fig_Frame}. Given an input with arbitrary modalities, MAT first presents a modality-adaptive feature extractor (MAFE) to extract discriminative unimodal features from each modality within the input. Here, MAFE takes an image of arbitrary modality along with its corresponding modality prompt as the input and leverages the modality prompt to adjust MAFE's feature space according to the characteristics of the input modality for extracting discriminative unimodal features.  Then, MAT designs a channel-wise and spatial-wise fusion hybrid (CSFH) strategy to dynamically fuse the unimodal features of all modalities at different levels for extracting complementary information across modalities. Specifically, considering that different levels of unimodal features have different properties, CSFH employs different fusion modules, \ie spatial-wise dynamic fusion module (SDFM) and channel-wise dynamic fusion module (CDFM), for different levels of unimodal features to effectively exploit cross-modal complementarity. Finally, those fuse features are fed into a saliency decoder to predict the final saliency maps.
Details about each component will be introduced in the following content.

\subsection{Modality-adaptive Feature Extractor (MAFE)}

The diverse discrepancies among multiple modalities usually result in challenges of extracting discriminative unimodal features from different modalities simultaneously by using one network or a few parameters. Inspired by prompt learning, our proposed modality-adaptive feature extractor (MAFE) tries to introduce a modality prompt for each modality during the feature extraction process to address the above issue. 
It will first leverage those modality prompts to adaptively adjust/tune its feature space according to the characteristics of the input modality. Then, in its training process, a novel modality translation contractive (MTC) loss is designed to learn those modality prompts to make them more modality-distinguishable when tuning feature spaces in the test stage. Finally, MAFE will be able to extract discriminative unimodal features from arbitrary modality with the aid of corresponding modality prompts. In the following content, we will first introduce the structure of MAFE and then detail the MTC loss.

\subsubsection{Network structure} As shown in Fig. \ref{fig_Frame}, our proposed MAFE takes an image $X_{M} \in \mathbb{R}^{W\times H \times C_1}$ of arbitrary modality with its corresponding modality prompt $P_{M} \in \mathbb{R}^{N_{mpt} \times C_2}$ as the inputs. Here, $M \in \{R, D, T\}$ denotes different modality data and $W$ and $H$ denote the width and height of the input image, respectively. $N_{mpt}$ denotes the length of modality prompts. $C_1$ and $C_2$ denote the numbers of channels. It first reshapes the image $X_{M} \in \mathbb{R}^{W\times H \times C_1}$ into a sequence of flattened 2D patches $X_{pa, M} \in \mathbb{R}^{K\times \frac{WH}{K}\times C_1}$, where $K$ denotes the number of total patches. Then, MAFE employs a liner projection layer to project $X_{pa, M}$ into patch embeddings $\mathbf{F}_{M}^{0} \in \mathbb{R}^{K\times C_2}$, \ie
\begin{equation}
	\mathbf{F}_{M}^{0} = \operatorname{Proj}\left(X_{pa, M};\alpha \right),
\end{equation}
where $\operatorname{Proj}\left(*;\alpha \right)$ denotes the linear projection layer and its parameters $\alpha$. 
After that, MAFE concatenates the patch embeddings $\mathbf{F}_{M}^{0}$ with the modality prompt $P_{M}$ and feeds them into a Transformer to extract multi-level features $\mathbf{F}_{M}^{1},..,\mathbf{F}_{M}^{L} $, \ie
\begin{equation}
	\mathbf{F}_{M}^{1},..,\mathbf{F}_{M}^{L} = \operatorname{Transformer}\left( \operatorname{Cat} \left(\mathbf{F}_{M}^{0}, P_{M} \right) \right).
\end{equation}
By utilizing Transformer's self/cross-attention mechanism, the modality prompts and unimodal features will significantly interact with each other in the feature extraction process, thus enabling the modality prompts to tune/adjust Transformer's feature space for accurately fitting the distributions of the input modalities.  

In this paper, MAFE is modified from the Pyramid Vision Transformer (PVTv2) \cite{PVTv2}. We slightly adjust PVTv2's linear spatial reduction attention module at the code level without changing its structure to support our modality prompts. It should be noted that other pyramid structure-based Transformer, such as Swin Transformer \cite{n31} and Twins Transformer \cite{TwinT}, can also be used. Evetually, four levels of unimodal features $\mathbf{F}_{M}^{1} \in \mathbb{R}^{\frac{W}{4}\times \frac{H}{4}\times 64}$, $\mathbf{F}_{M}^{2} \in \mathbb{R}^{\frac{W}{8}\times \frac{H}{8}\times 128}$, $\mathbf{F}_{M}^{3} \in \mathbb{R}^{\frac{W}{16}\times \frac{H}{16}\times 320}$ and $\mathbf{F}_{M}^{4}\in \mathbb{R}^{\frac{W}{32}\times \frac{H}{32}\times 512}$ are obtained. Besides, before feeding those images of different modalities into MAFE, we first unify their channels to 3 by the replication operation.

\subsubsection{Modality Translation Contractive (MTC) loss} MTC loss is designed to further optimize the training process for learning more modality-distinguishable modality prompts, enhancing their abilities to tune/adjust feature space according to the characteristics of the input modalities in the testing stage. 
﻿
\begin{figure}[!t]
	\centering
	\includegraphics[width=\linewidth]{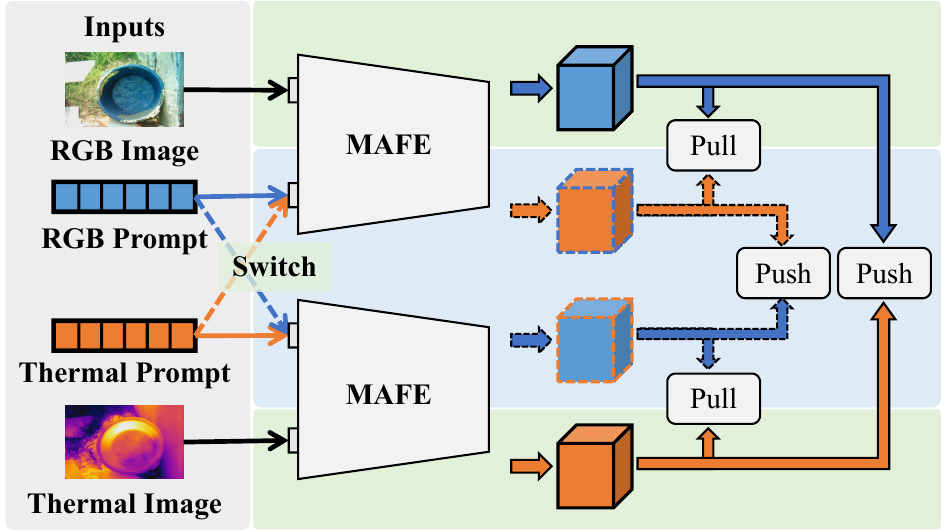}
	\caption{Diagram of our proposed MTC loss. Take the RGB images and thermal images as the example. The features extracted from an RGB image with an RGB prompt should have different distributions with the features extracted from a thermal image with  a thermal prompt, but share similar distributions with the features extracted from a thermal image with  a thermal prompt.}
	\label{fig_rec_loss}
\end{figure}

As shown in Fig. \ref{fig_rec_loss}, the design motivation of our proposed MTC loss is that the ideal MAFE should have the ability to fit the distribution of different modalities with the aid of different modality prompts. Accordingly, MAFE will exhibit two different behaviors. First, if feeding an image of a specific modality with its corresponding modality prompt, the features extracted by MAFE should be able to well represent the characteristics of this modality data.  Secondly, if feeding an image of a specific modality with the modality prompt of another modality, the features extracted by MAFE  should follow the same characteristics with the second modality (\ie the modality corresponding to the modality prompt), since MAFE will adjust its feature space according to the modality prompt. Such two behaviors will lead to that the features extracted from the images with the same modality prompt should share similar distributions, while the features extracted from the images with different modality prompts should have different  distributions considering modality discrepancies.
Based on such motivation, MTC loss is conducted by the following steps.
	
Suppose that there are an image $X_{M_1}$ of $M_1$ modality, an image $X_{M_2}$ of $M_2$ modality and their modality prompts $P_{M_1}$ and $P_{M_2}$.	 Here, the images  $X_{M_1}$ and $X_{M_2}$ are completely  registered to each other. MAFE will first take ($X_{M_1}$, $P_{M_1}$ ) and ($X_{M_2}$, $P_{M_2}$) as the inputs, respectively, and accordingly extract four levels of unimodal features $\mathbf{F}_{M}^{1},...,\mathbf{F}_{M}^{4}$. Here, $M \in \{M_1, M_2\}$ denotes different modalities.  Then, MAFE will further change their modality prompts for extracting another four levels of unimodal features $\mathbf{\hat{F}}_{M}^{1},...,\mathbf{\hat{F}}_{M}^{4}$, \ie MAFE takes ($X_{M_1}$, $P_{M_2}$ ) and ($X_{M_2}$, $P_{M_1}$) as the inputs. 
After that, our proposed MTC loss will pull the features extracted from the images with the same modality prompts closer to each other and push the features extracted from the images with different modality prompts further away from each other by 
\begin{equation}
\mathcal{L}_{MTC} = \sum_{l=1}^{4}\left(
\frac{\operatorname{exp}\left(\operatorname{Ds}(\mathbf{F}_{M_1}^{l}, \mathbf{\hat{F}}_{M_2}^{l}) +\operatorname{Ds}(\mathbf{\hat{F}}_{M_1}^{l}, \mathbf{F}_{M_2}^{l})\right)}{\operatorname{exp}\left(\operatorname{Ds}(\mathbf{F}_{M_1}^{l}, \mathbf{F}_{M_2}^{l}) +\operatorname{Ds}(\mathbf{\hat{F}}_{M_1}^{l}, \mathbf{\hat{F}}_{M_2}^{l})\right)}
\right),
\end{equation}
where $\operatorname{exp}(*)$ denotes the exponential function. $\operatorname{Ds}(*)$ denotes the distance function, such as euclidean distance.

By virtue of MTC loss, our proposed MAFE can learn those modality-distinguishable modality prompts for effectively adjusting the feature space of MAFE according to the characteristics of the input modalities, thus being able to capture discriminative unimodal features from arbitrary modality and address diverse discrepancies among multiple modalities.

\subsection{Channel-wise and Spatial-wise Fusion Hybrid (CSFH) strategy}

After obtaining those unimodal features, the next step is to exploit their cross-modal complementary information for boosting performance. For that, a novel channel-wise and spatial-wise fusion hybrid (CSFH) strategy is designed to effectively fuse multiple types of unimodal features extracted from an arbitrary number of modalities. CSFH first designs a spatial-wise dynamic fusion module (SDFM) and a channel-wise dynamic fusion module (CDFM) to explore the spatial-wise and channel-wise relations among different modalities for exploiting their complementary semantic and detail information. Then, according to the properties of different levels of unimodal features, CSFH carefully aligns the SDFM and CDFM to the features of different levels for capturing more complementary information. In the following content, we will first introduce the two dynamic fusion modules and then introduce our hybrid strategy. 

\begin{figure}[!t]
	\centering
	\includegraphics[width=\linewidth]{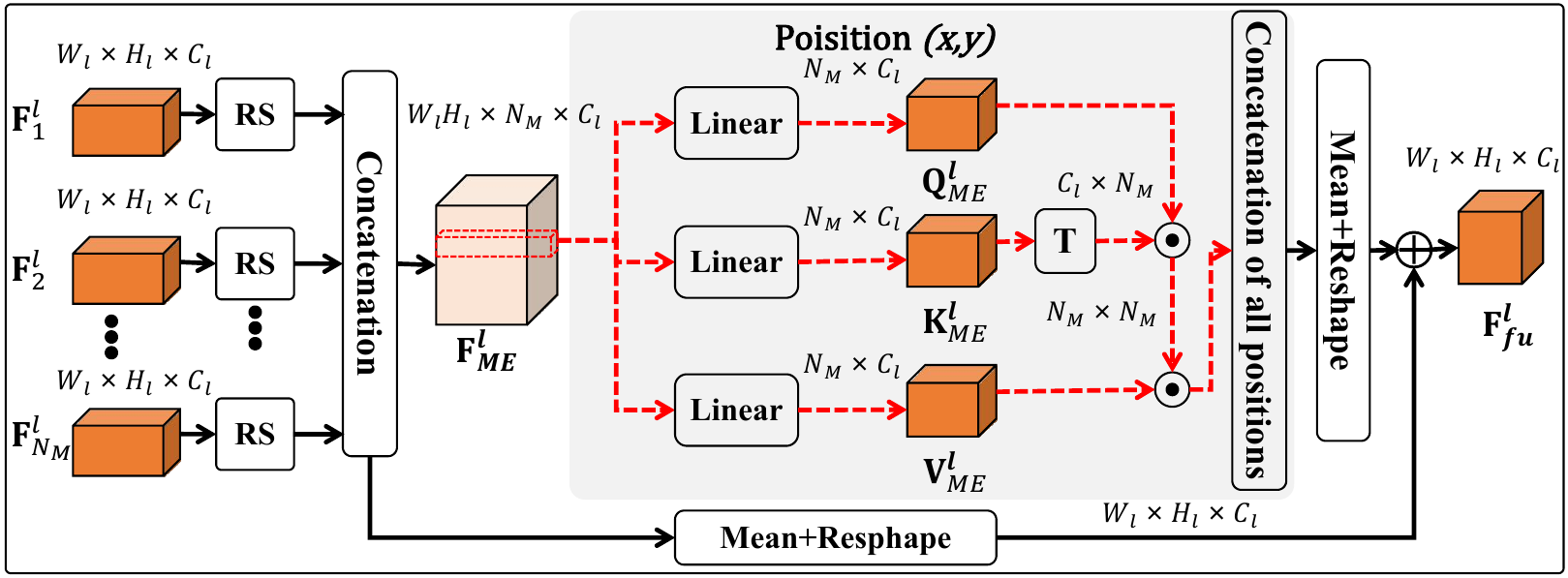}
	\caption{Architecture of our proposed SDFM. }
	\label{fig_SDFM}
\end{figure}

\subsubsection{Spatial-wise dynamic fusion module (SDFM)} SDFM aims to dynamically fuse those unimodal features from an arbitrary number and effectively capture their cross-modal complementary detail information by exploring the spatial-wise interactions among modalities with arbitrary numbers. 
Suppose that the inputs contain the unimodal features extracted from $N_M$ modalities ($N_M>1$) and their corresponding unimodal features are denoted by $\{ \mathbf{F}_{n_m}^{l} \in \mathbb{R}^{W_l \times H_l\times C_l} \}_{n_m=1}^{N_M}$. Here, $n_m = 1,..., N_M$ denotes different modalities and $l=1,2,..,4$ denotes different feature levels. $W_l$, $H_l$ and $C_l$ denote the width, height and channels of the features $\mathbf{F}_{n_m}^{l} $, respectively. The structure of SDFM is shown in Fig. \ref{fig_SDFM}, which has the following steps.

First, SDFM will conduct modality token embedding $\mathbf{F}_{ME}^{l} \in \mathbb{R}^{W_lH_l \times N_M\times  C_l}$ by reshaping the unimodal features' size of each modality from  $W_l \times H_l\times C_l$ into $W_lH_l \times 1\times  C_l$ and concatenating those reshaped unimodal features of $N_M$ modalities, \ie
\begin{equation}
	\mathbf{F}_{ME}^{l} = \operatorname{Cat}\left( \operatorname{RS}(\mathbf{F}_{1}^{l}), \operatorname{RS}(\mathbf{F}_{2}^{l}),..., \operatorname{RS}(\mathbf{F}_{N_M}^{l}) \right),
\end{equation}
where $\operatorname{Cat}\left( * \right)$ denotes the concatenation operation. $\operatorname{RS}(*)$ denotes the reshape operation.

Then, it will explore spatial-wise relations among modalities within modality token embedding $\mathbf{F}_{ME}^{l}$ by employing the self-attention mechanism to establish interactions among each spatial location $(x, y)$. For that, the key features $\mathbf{K}_{ME}^{l}(x, y)$, query features $\mathbf{Q}_{ME}^{l}(x, y)$ and value features $\mathbf{V}_{ME}^{l}(x, y) \in \mathbb{R}^{N_M \times C_l}$ are first obtained by using different linear projection functions $\operatorname{Proj}\left(*; \zeta_{sk} \right)$, $\operatorname{Proj}\left(*; \zeta_{sq} \right)$ and $\operatorname{Proj}\left(*; \zeta_{sv} \right)$, \ie
\begin{equation}
	\begin{split}
		& \mathbf{K}_{ME}^{l}(x, y) = \operatorname{Proj}\left(\mathbf{F}_{ME}^{l}(x, y); \zeta_{sk} \right) , \\ 
		& \mathbf{Q}_{ME}^{l}(x, y) = \operatorname{Proj}\left(\mathbf{F}_{ME}^{l}(x, y); \zeta_{sq} \right) , \\ 
		& \mathbf{V}_{ME}^{l}(x, y) = \operatorname{Proj}\left(\mathbf{F}_{ME}^{l}(x, y); \zeta_{sv} \right) , 
	\end{split}
\end{equation}
where $\zeta_{sk}$, $\zeta_{sq}$, $\zeta_{sv}$ denote corresponding parameters. Then, their interaction weights $W_{ME}^{l}(x,y)$ are obtained by
\begin{equation}
	W_{ME}^{l}(x,y) = \operatorname{Softmax}\left(\frac{\mathbf{Q}_{ME}^{l}(x, y)(\mathbf{K}_{ME}^{l}(x, y))^T}{\sqrt{C_l}}\right), \\ 
\end{equation}
where $\operatorname{Softmax}(*)$ denotes softmax function and $(*)^T$ denotes the transposition operation. The weights $W_{ME}^{l}(x,y)$ explore the spatial-wise relations among the position $(x,y)$ of different modalities.  Finally, based on $W_{ME}^{l}(x,y)$, the spatial-wise interactions among different modalities are established by
\begin{equation}
	\mathbf{F}_{It}^{l}(x,y)= W_{ME}^{l}(x,y)\mathbf{V}_{ME}^{l}(x, y).
\end{equation}
﻿
﻿
After that, SDFM concatenates the interacted features $\mathbf{F}_{It}^{l}(x,y)$ of all positions and obtains their preliminary fused features $\mathbf{\hat{F}}_{fu}^{l} \in \mathbb{R}^{W_l \times H_l\times C_l}$ by using the mean opeartion, \ie 
\begin{equation}
	\mathbf{\hat{F}}_{fu}^{l} = \operatorname{RS}\left( \operatorname{Mean}\left( \operatorname{Cat}\left(\mathbf{F}_{It}^{l}(1,1),..., \mathbf{F}_{It}^{l}(W_l,H_l)  \right) \right) \right).
\end{equation}
﻿
Finally, SDFM will obtain the final fused features $\mathbf{F}_{fu}^{l} \in \mathbb{R}^{W_l \times H_l\times C_l}$  by adding the feed forward network $\operatorname{FFN}(*; \nu)$, \ie
\begin{equation}
	\mathbf{F}_{fu}^{l} =  \operatorname{FFN}(\operatorname{Mean}\left( \mathbf{F}_{1}^{l}, \mathbf{F}_{12}^{l}, ..., \mathbf{F}_{N_M}^{l}\right); \nu ) + \mathbf{\hat{F}}_{fu}^{l},
\end{equation}
where $\nu$ denotes the parameters. Besides, if $N_M=1$, \ie there is only one modality in the input, the final fused features $\mathbf{F}_{fu}^{l}$ will be obtained by
\begin{equation}
	\mathbf{F}_{fu}^{l} =  \operatorname{FFN}(\mathbf{F}_{1}^{l}; \nu)  + \mathbf{V}_{ME}^{l}.
\end{equation}
By doing so, SDFM can establish the spatial-wise interactions among modalities at position $(x,y)$, thus effectively capturing their complementary detail information.

\begin{figure}[!t]
	\centering
	\includegraphics[width=\linewidth]{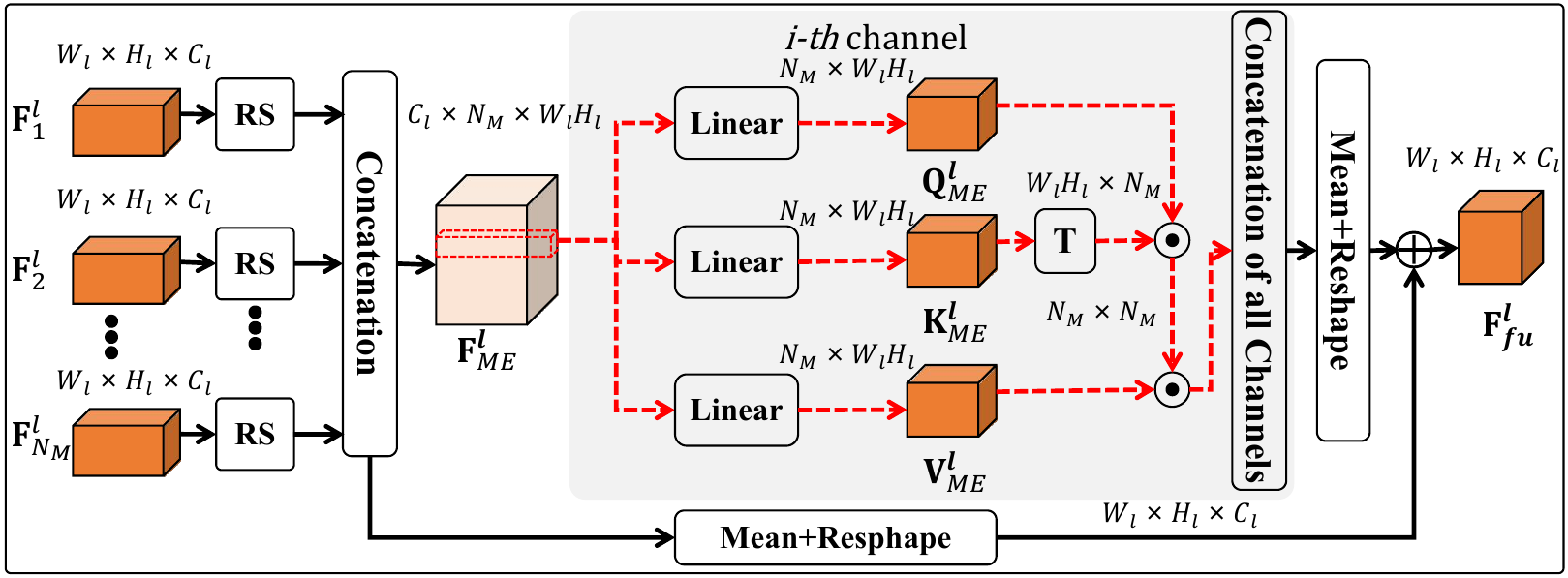}
	\caption{Architecture of our proposed CDFM. }
	\label{fig_CDFM}
\end{figure}

\subsubsection{Channel-wise dynamic fusion module (CDFM)} The proposed CDFM follows a similar structure to the DFM proposed in [] with slight modification.  
﻿
Specifically, as shown in Fig. \ref{fig_CDFM}, similar to SDFM, CDFM also  first conducts the modality token embeddings  
$\mathbf{F}_{ME}^{l} \in \mathbb{R}^ {C_l \times N_M\times  W_lH_l}$ from its inputs $\{ \mathbf{F}_{n_m}^{l} \}_{n_m=1}^{N_M}$. Then, instead of exploring spatial-wise relations among modalities, CDFM explores channel-wise interactions among modalities  by employing the self-attention mechanism for each channel of the modality token embeddings  
$\mathbf{F}_{ME}^{l}$ rather than the position $(x,y)$. After that, CDFM further uses the mean operation to fuse those interacted features for extracting more complementary semantic information, thus obtaining preliminary fused features $\mathbf{\bar{F}}_{fu}^{l}$. Finally, the final fused features $\mathbf{F}_{fu}^{l}$ are also obtained by using the feed forward network.  

\subsubsection{Hybrid strategy} Instead of directly employing SDFM and CDFM for each level of those extracted unimodal features, CSFH aligns them for different levels of unimodal features in a hybrid strategy.

Specifically, many existing works have proved that pyramid structure-based Transformers, such as PVT \cite{PVTv2} and Swin Transformers \cite{n31}, focus on different characteristics of the inputs in their different levels of features. Generally speaking, low-level features (or Tokens) may pay more attention to local dependences and tend to extract those detail/texture information. While, high-level features may concentrate on those global dependences and usually capture those global semantic information.

Therefore, CSFH employs the proposed SDFM to fuse those low-level unimodal features $\{ \mathbf{F}_{n_m}^{l} \}_{n_m=1}^{N_M}$($l=1,2$) for extracting their complementary detail information, \ie
\begin{equation}
	\mathbf{F}_{fu}^{l} = \operatorname{SDFM}\left( \mathbf{F}_{1}^{l}, \mathbf{F}_{2}^{l},...,\mathbf{F}_{N_M}^{l}; \gamma_{sdfm}  \right).
\end{equation}
Here, $l=1,2$ denotes the features in low levels. $\operatorname{SDFM}\left( *; \gamma_{sdfm}  \right)$ denotes our proposed SDFM module with its parameters $ \gamma_{sdfm} $.  Correspondingly, for high-level features $\{ \mathbf{F}_{n_m}^{l} \}_{n_m=1}^{N_M}$($l=3,4$),  CSFH employs the proposed CDFM to fuse them for extracting their complementary detail information, \ie
\begin{equation}
	\mathbf{F}_{fu}^{l} = \operatorname{CDFM}\left( \mathbf{F}_{1}^{l}, \mathbf{F}_{2}^{l},...,\mathbf{F}_{N_M}^{l}; \gamma_{cdfm}  \right),
\end{equation}
where $\operatorname{CDFM}\left( *; \gamma_{cdfm}  \right)$ denotes our proposed CDFM module with its parameters $ \gamma_{cdfm} $. 

\subsection{Decoder}
﻿
After obtaining multi-level of fused features, the next step is to detect those salient objects from them. As shown in Fig. \ref{fig_Frame}, the saliency decoder of our proposed model follows the classic coarse-to-fine network structure. Specifically, the $i$-th level features will be first fused with its higher level features (if existed), and then will be fed into their lower level features for further fusion. Finally, the saliency maps $\mathbf{S}^1$ will be predicted from the last level features.  Moreover, inspired by multi-level supervision strategy, three auxiliary saliency maps  $\mathbf{S}^2$, $\mathbf{S}^3$ and $\mathbf{S}^4$ will be also deduced from other levels of features in the decoding processing for accelerating training process.   

\subsection{Loss Function}

Following \cite{n51} and \cite{n50}, the cross-entropy loss $\mathcal{L}_{CE} $ and the edge loss $\mathcal{L}_{Edge} $ are jointly employed for optimizing our proposed model. According, their combined loss  $\mathcal{L}_{Sal}$ is expressed by
\begin{equation}
	\mathcal{L}_{Sal} =  \sum_{l=1}^{4} \left( \mathcal{L}_{CE}(\mathbf{S}^l, \mathbf{S}^{l}_{gt}) + \mathcal{L}_{Edge}(\mathbf{S}^l, \mathbf{S}^{l}_{gt}) \right),
\end{equation} 
where $\mathbf{S}^{l}_{gt}$ denotes the ground truth saliency maps of the $l$-th level. Here, the cross-entropy loss $\mathcal{L}_{CE} $  is expressed by
\begin{equation}
	\mathcal{L}_{CE}(\mathbf{Y}, \mathbf{\hat{Y}})=  \mathbf{Y}\log (\mathbf{\hat{Y}})+(1-\mathbf{Y})\log (1-\mathbf{\hat{Y}}),
\end{equation}
where $\mathbf{\hat{Y}}$ and $\mathbf{Y}$ denote the predicted value and its corresponding ground truth value, respectively. And, the edge loss $\mathcal{L}_{Edge} $  are expressed by
\begin{equation}
	\mathcal{L}_{Edge}(\mathbf{Y}, \mathbf{\hat{Y}})=  \operatorname{Mse}\left( \operatorname{Sobel}\left( \mathbf{Y} \right), \operatorname{Sobel}\left( \mathbf{\hat{Y}} \right) \right),
\end{equation}
where $\operatorname{Sobel}(*)$ denotes the Sobel edge detection operation. 

Accordingly, the total loss for training  our proposed model is formulated as:
\begin{equation}
	\mathcal{L}_{total}= \mathcal{L}_{Sal} + \mathcal{L}_{MTC}.
\end{equation} 

\section{Experiments} \label{sec::ex}

\subsection{Datasets and Evaluation Metrics}

\subsubsection{Datasets} We train and test our proposed MAT by using a recently proposed AM-XD dataset.  Its training set comprises a total of 11533 samples, encompassing 5000 RGB SOD images, 2985 RGB-D SOD image pairs, 2500 RGB-T SOD image pairs, and 1048 RGB-D-T image pairs, respectively. Its testing set, on the other hand, comprises a total of 13442 samples, which consist of 5000 RGB images, 3121 RGB-D image pairs, 4321 RGB-T image pairs, and 1000 RGB-D-T image pairs, respectively. There are two testing modes of AM-XD dataset, \ie the  sole mode and the joint mode. The former assesses SOD models across seven distinct subsets, \ie RGB SOD testing set, D SOD testing set, T SOD testing set, RGB-D SOD testing set, RGB-T SOD testing set, D-T SOD testing set, and RGB-D-T SOD testing set, respectively. The latter does not distinguish different subsets but tests different models by jointly using all samples.

\begin{table*}[!t]
	\renewcommand{\arraystretch}{1.3}
	\caption{Quantitative results of different models on SOD dataset.} 
	\label{tab_88}
	\centering
	\resizebox{\textwidth}{!}{
		\begin{tabular}{cc|cccccc|cccccc|cc|cc}
			\hline 
			& & \multicolumn{14}{c|}{Sole Mode} & \multicolumn{2}{c}{Joint Mode} \\
			\hline
			& & \multicolumn{6}{c|}{Single-modality SOD} & \multicolumn{6}{c|}{Two-modality SOD}& \multicolumn{2}{c|}{Three-modality SOD}& \multicolumn{2}{c}{ } \\
			\hline 
			& \multirow{2}{*}{ Models } & \multicolumn{2}{c|}{RGB} & \multicolumn{2}{c|}{D}  & \multicolumn{2}{c|}{T} & \multicolumn{2}{c|}{RGB-D} & \multicolumn{2}{c|}{RGB-T}& \multicolumn{2}{c|}{D-T}& \multicolumn{2}{c|}{RGB-D-T} & \multicolumn{2}{c}{ALL} \\
			\cline{3-18}
			&  &$\mathcal{M}\downarrow$ &$F_\beta\uparrow$ &$\mathcal{M}\downarrow$ &$F_\beta\uparrow$ &$\mathcal{M}\downarrow$ &$F_\beta\uparrow$ &$\mathcal{M}\downarrow$ &$F_\beta\uparrow$ &$\mathcal{M}\downarrow$ &$F_\beta\uparrow$ &$\mathcal{M}\downarrow$ &$F_\beta\uparrow$ &$\mathcal{M}\downarrow$ &$F_\beta\uparrow$&$\mathcal{M}\downarrow$ &$F_\beta\uparrow$ \\
			\hline 
			\multirow{3}{*}{\shortstack{RGB \\ SOD}}&  PSGLoss(2021) \cite{o14}&0.047&0.722&0.099&0.482&0.070&0.507 &- & - &- &- &-& -&- &- &- &-  \\ 
			&  PoolNet++(2023) \cite{o13} &0.041&0.822&0.120&0.450&0.810&0.458&- & - &- &- &-& -&- &- &- &-  \\ 
			&  SefReFormer(2023) \cite{o15}& \textbf{0.033} &{0.844}    & 0.098    
			&0.576 &0.066    & 0.717 &- & - &- &- &-& -&- &-&- &-  \\  
			\hline  
			\multirow{3}{*}{\shortstack{RGB-D \\ SOD}} & VST(2022)\cite{n33} &0.044&0.817&0.093&0.629&0.083&0.682 &0.038&0.816&0.035&0.803&0.074&0.493&- &- &- &-  \\ 
			&  SwinNet(2022) \cite{n30} &- &- &- &- &- &- &0.045&0.786&{0.028}&{0.837}&0.027&0.518&- &- &- &- \\
			& CAVER(2023)\cite{z27}  &- &- &- &- &- &- &0.044&0.715&0.038&0.76&0.089&0.469&- &- &- &-\\
			\hline 
			\multirow{3}{*}{\shortstack{RGB-T \\ SOD}}&  APNet(2022) \cite{z14}&- &- &- &- &- &- &0.066&0.695&0.036&0.743&0.036&0.422&- &- &- &-    \\
			& FANet(2023)\cite{z29} &- &- &- &- &- &- &0.067&0.712&0.035&0.767&0.061&0.382&- &- &- &-  
			\\
			& LSNet(2023)\cite{z25} &- &- &- &- &- &- &0.064&0.714&0.060&0.676&0.064&0.390&- &- &- &-  \\	
			\hline  
			\multirow{2}{*}{\shortstack{RGB-D-T \\ SOD}} & HWSI (2023)\cite{9931143} &- &- &- &- &- &- &- &- &- &- &- &-&\textbf{0.0026} & \textbf{0.896} &- &-    \\
			& MFFNet (2023)\cite{10171982} &- &- &- &- &- &- &- &- &- &- &- &-&0.0032 &  0.871 &- &-   \\ 
			\hline 
			\multirow{2}{*}{\shortstack{AM \\ SOD}}  &  MSN & 0.055  &0.803 &{0.064} &{0.700} & 0.044 & 0.769  &  0.035 &0.850 &0.038 &0.833 & 0.0065& 0.740 &0.0035 &  0.841  &0.049     &0.816\\ 
			& OUR&0.035&\textbf{0.845}&\textbf{0.058}&\textbf{0.720}&\textbf{0.036}&\textbf{0.799} &\textbf{0.029}&\textbf{0.865}&\textbf{0.023}&\textbf{0.860}&\textbf{0.0052}&\textbf{0.774}&0.0035&0.845&\textbf{0.033}&\textbf{0.854} \\ 
			\hline
		\end{tabular}
	}  
\end{table*}

\begin{figure*}[!t]
	\centering
	\includegraphics[width=0.75\linewidth]{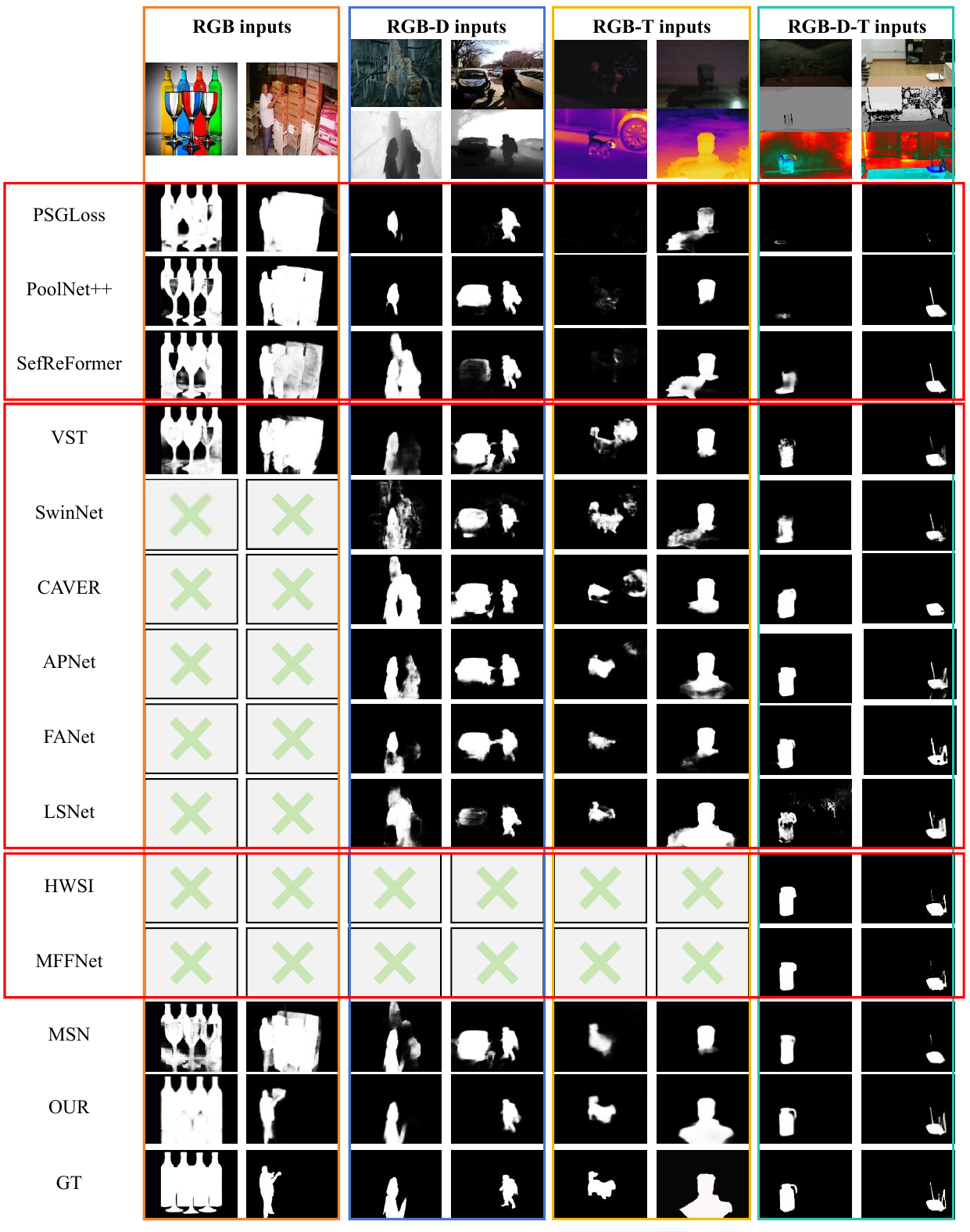}
	\caption{Visualization results of different models. }
	\label{fig_com}
\end{figure*}

\subsubsection{Evaluation Metrics}

Two widely used metrics, \ie mean absolute error ($\mathcal{M}$)\cite{NJU} and mean F-measure ($F_\beta$)\cite{NJU}, are employed to evaluate the performance of different models. 
Here, the mean absolute error ($\mathcal{M}$) quantifies the average absolute deviation between the predicted saliency map $\mathbf{S}$ and the ground truth $\mathbf{S}_{gt}$ by 
\begin{equation}\label{eq_mae}
	\mathcal{M} = \frac{1}{W \times H} \sum_{x=1}^{W} \sum_{y=1}^{H}|\mathbf{S}(x,y)-\mathbf{S}_{gt}(x,y)|,
\end{equation}
where $W$ and $H$ denote the width and height of the saliency map (or ground truth), respectively. $|*|$ denotes the operation of computing absolute values.
Meanwhile, the F-measure ($F_\beta$)  offers a comprehensive measure of accuracy by harmonizing precision and recall into a unified score, \ie
\begin{equation}\label{eq_fm}
	F_{\beta} = \frac{(1+\omega^2) \times Precision \times Recall}{\omega^2 \times Precision + Recall}.
\end{equation}
Here, $Precision$ calculates the fraction of true positive predictions among all positive predictions, while  $Recall$ represents the fraction of true positive predictions among all actual positive instances.  Consistent with \cite{NJU}, we set $\omega^2$ to 0.3.

\subsection{Implementation details}

We utilize the popular Pytorch library to compile our proposed model and execute it on an NVIDIA 3090Ti GPU. Leveraging the benefits of pre-trained models, we initialize the parameters of our MAFE model based on a PVTv2 network which has been pre-trained on the ImageNet dataset\cite{n10}. We adopt the Kaiming initialization method\cite{KaiMingInit} for initializing the parameters of other modules.    In the training stage, we employ the SGD algorithm\cite{SGD} with Nesterov momentum for optimization, setting the learning rate to 2e-3 and the weight decay to 5e-4. We first jointly train the total network of about 60 epochs and then only fine-tune the modality prompts about 5 epochs.  The sizes of all inputs are resized into $224 \times 224$.

\subsection{Quantitative comparisons with SOTA models}

As shown in Table \ref{tab_88}, we conduct a comparative analysis of our proposed AM model with several state-of-the-art SOD models, including \textbf{RGB SOD models}(PSGLoss \cite{o14}, PoolNet++ \cite{o13} and SefReFormer \cite{o15}), \textbf{RGB-D SOD models} (VST(2022)\cite{n33},  SwinNet(2022) \cite{n30},  CAVER(2023)\cite{z27}), \textbf{RGB-T SOD models} (APNet(2022) \cite{z14},  FANet(2023)\cite{z29} and LSNet(2023)\cite{z25}), and \textbf{RGB-D-T SOD models}(HWSI(2023)\cite{9931143} and MFFNet(2023)\cite{10171982}). 
It's worth noting that those RGB-T SOD and RGB-D SOD models actually  train their models on both RGB-T SOD and RGB-D SOD datasets, respectively, \ie these models actually perform two-modality salient object detection. Furthermore, VST \cite{n33}  also has an RGB SOD version, and its corresponding RGB SOD results are provided as well.

It can be seen that, for sole mode, our proposed MAT achieves the best performance under the settings of D SOD, T SOD, RGB-D SOD, RGB-T SOD and D-T SOD, respectively.  While our proposed MAT obtains competitive and even better results under the RGB SOD setting.  However, our proposed model obtains suboptimal results for the RGB-D-T setting. While, for the joint mode, our proposed model significantly improves the performance of AM-XD dataset than existing AM SOD models. This mainly results from that our proposed MAT can adaptively adjust its feature space according to the characteristics of input modality by using those learned modality prompts, thus enabling it to effectively represent each modality in a large feature space. Meanwhile, our proposed CSFH can effectively capture cross-modal complementary information by simultaneously exploring channel-wise and spatial-wise relations across modalities.

\subsection{Qualitative comparisons with SOTA models}

The visualization results of different models are also shown in Fig. \ref{fig_com}. Considering that RGB SOD, RGB-D/RGB-T SOD, and RGB-D-T SOD are widely studied tasks, we compare our proposed model with existing models on those settings for fair comparison and better exhibits within limited space. Specifically, for RGB SOD models (\ie, PSGLoss \cite{o14}, PoolNet++ \cite{o13}, and SefReFormer \cite{o15}), we visualize their corresponding results for RGB images in the settings of RGB, RGB-D/RGB-T, and RGB-D-T inputs. For RGB-D/RGB-T SOD models (\ie, VST(2022)\cite{n33}, SwinNet(2022) \cite{n30}, CAVER(2023)\cite{z27}, APNet(2022) \cite{z14}, FANet(2023)\cite{z29}, and LSNet(2023)\cite{z25}), we visualize their results for RGB-D images with RGB-D inputs and for RGB-T images with RGB-T inputs, since those models are independently trained under RGB-D and RGB-T settings. For RGB-D-T inputs, we visualize the results of RGB-D SOD models for RGB-D inputs and the results of RGB-T SOD models for RGB-T inputs.

\begin{table*}[!t]
	\renewcommand{\arraystretch}{1.3}
	\caption{Quantitative results of each component of our proposed model.} 
	\label{tab_98}
	\centering
	\resizebox{\textwidth}{!}{
		\begin{tabular}{c|c|cc|cc|cc|cc|cc|cc|cc|cc}
			\hline 
			\multirow{3}{*}{Models}	& \multirow{2}{*}{Params}& \multicolumn{14}{c|}{Sole} &   \multicolumn{2}{c}{Joint} \\		\cline{3-18} 
			& & \multicolumn{2}{c|}{RGB} & \multicolumn{2}{c|}{D}  & \multicolumn{2}{c|}{T} & \multicolumn{2}{c|}{RGB-D} & \multicolumn{2}{c|}{RGB-T}& \multicolumn{2}{c|}{D-T}& \multicolumn{2}{c|}{RGB-D-T} & \multicolumn{2}{ c}{ALL} \\
			\cline{2-18} 
			&M &$\mathcal{M}\downarrow$ &$F_\beta\uparrow$ &$\mathcal{M}\downarrow$ &$F_\beta\uparrow$ &$\mathcal{M}\downarrow$ &$F_\beta\uparrow$ &$\mathcal{M}\downarrow$ &$F_\beta\uparrow$ &$\mathcal{M}\downarrow$ &$F_\beta\uparrow$ &$\mathcal{M}\downarrow$ &$F_\beta\uparrow$ &$\mathcal{M}\downarrow$ &$F_\beta\uparrow$&$\mathcal{M}\downarrow$ &$F_\beta\uparrow$ \\
			\hline 
			Baseline &37.76&0.045&0.815&0.071&0.688&0.046&0.778&0.034&0.843&0.028&0.841&0.0062&0.744&0.0049&0.789&0.041&0.830 \\
			+Modality Prompts	& 37.76	&0.039&0.831&0.065&0.699&0.041&0.786&0.031&0.856&0.026&0.85&0.0055&0.764&0.0044&0.805&0.036&0.845\\	
			+Modality Prompts+ $ \mathcal{L}_{MTC}$ (MAFE)	& 37.76&0.037&0.837&0.063&0.707&0.039&0.793&0.030&0.860&0.025&0.854&0.0054&0.769&0.0036 &  0.833&0.035&0.848
			\\  
			+Modality Prompts+ $ \mathcal{L}_{MTC}$ + CSFH 	& \textbf{42.24}&\textbf{0.035}&\textbf{0.845}&\textbf{0.058}&\textbf{0.720}&\textbf{0.036}&\textbf{0.799}&\textbf{0.029}&\textbf{0.865}&\textbf{0.023}&\textbf{0.860}&\textbf{0.0052}&\textbf{0.774}&\textbf{0.0035}&\textbf{0.845}&\textbf{0.033}&\textbf{0.854}  \\  
			\hline
	\end{tabular}} 
\end{table*}

\begin{table*}[!t]
	\renewcommand{\arraystretch}{1.3}
	\caption{Quantitative results of each component of our proposed model.} 
	\label{tab_998}
	\centering
	\resizebox{\textwidth}{!}{
		\begin{tabular}{c|c|cc|cc|cc|cc|cc|cc|cc|cc}
			\hline 
			\multirow{3}{*}{SDFM Levels}	& \multirow{3}{*}{CDFM Levels}& \multicolumn{14}{c|}{Sole} &   \multicolumn{2}{c}{Joint} \\		\cline{3-18} 
			& & \multicolumn{2}{c|}{RGB} & \multicolumn{2}{c|}{D}  & \multicolumn{2}{c|}{T} & \multicolumn{2}{c|}{RGB-D} & \multicolumn{2}{c|}{RGB-T}& \multicolumn{2}{c|}{D-T}& \multicolumn{2}{c|}{RGB-D-T} & \multicolumn{2}{ c}{ALL} \\
			\cline{3-18} 
			& &$\mathcal{M}\downarrow$ &$F_\beta\uparrow$ &$\mathcal{M}\downarrow$ &$F_\beta\uparrow$ &$\mathcal{M}\downarrow$ &$F_\beta\uparrow$ &$\mathcal{M}\downarrow$ &$F_\beta\uparrow$ &$\mathcal{M}\downarrow$ &$F_\beta\uparrow$ &$\mathcal{M}\downarrow$ &$F_\beta\uparrow$ &$\mathcal{M}\downarrow$ &$F_\beta\uparrow$&$\mathcal{M}\downarrow$ &$F_\beta\uparrow$ \\
			\hline 
			1,2,3,4  & -    &0.037         &0.841         &0.061         &0.713      &0.043&0.789&0.030&0.863&0.026&0.855&0.0052&0.775&0.0041&0.817&0.035&0.851 \\
			1,2,3	 & 4	&0.036         &0.842         &0.060         &0.713      &0.039&0.795&\textbf{0.029}&0.863&0.024&0.859&0.0051&0.775&0.0036&0.837  &\textbf{0.033}&0.854\\
			1, 2 	& 3, 4  &\textbf{0.035}&\textbf{0.845}&\textbf{0.058} &0.720&\textbf{0.036}&\textbf{0.799}&\textbf{0.029}&0.865&\textbf{0.023}&\textbf{0.860}&0.0052&0.774&\textbf{0.0035}&\textbf{0.845}&\textbf{0.033}&\textbf{0.854} \\  
			1	& 2,3,4     &\textbf{0.035}&0.844         &\textbf{0.058} &\textbf{0.721}&0.040&0.794&\textbf{0.029}&\textbf{0.866}&0.025&0.857&\textbf{0.0050}&\textbf{0.782}&0.0036 &0.835&\textbf{0.033}&0.854  \\  
			-	& 1, 2,3,4  &0.038         &0.839         &0.064         &0.707&0.044&0.785&0.030&0.862&0.027&0.852&0.0053&0.771&0.0038  &0.826 &0.036&0.849  \\  
			\hline
	\end{tabular}}
\end{table*}

It can be seen that, compared with existing models which can only detect salient objects from the inputs with a fixed number of modalities, AM SOD models can simultaneously detect salient objects from the inputs with arbitrary modalities. The first two columns show the RGB inputs with large salient objects (1$st$ column) or complex backgrounds (2$nd$ column),  existing RGB SOD models may only detect partial salient objects or mistakenly identify the backgrounds as the salient objects, while our proposed model obtains more complete and accurate saliency maps. This indicates that our proposed MAFE can effectively extract those discriminative unimodal features by virtue of corresponding modality prompts.  

The results in the third column to the eighth column indicate that using more modalities usually obtains better results than using fewer modalities. For example, the salient objects share similar hues with the background in the RGB images in the third column, and the RGB images cannot well capture scene information in the fifth and sixth columns. In those conditions, those two-modality SOD models achieve better results than RGB SOD models. Moreover, our proposed model obtains better results than those two-modality SOD models and existing AM SOD models, which may be due to the fact that the proposed CSFH strategy can effectively capture cross-modal complementary information by exploring the channel-wise or spatial-wise interactions across the unimodal features at different levels. Besides, it can also be seen that using more modalities may be redundant or even counterproductive in some special cases. For instance, the scene in the fourth column is captured under good light conditions, where there is a lot of redundant information across the RGB images and depth images, thus leading to closer results for the RGB SOD models and two-modality SOD models. However, the depth image in the eighth column tends to be of low quality, thus the results of those RGB-D SOD models may be inferior to those of RGB SOD models. Moreover, in those special cases, our proposed model can still achieve the best results, which further proves the effectiveness of our proposed model.

\subsection{Ablation study}

\subsubsection{Effectiveness of each component of our proposed model}

In this subsection, we conduct several experiments to verify the effectiveness of each component of our proposed model. Specifically, we first build the baseline model by removing the modality prompts from our proposed MAFE and replacing the proposed CSFH with the element-wise addition fusion way. Then, we gradually add different modules to the baseline, including modality prompts, MTC loss and CSFH. Their evaluation results are shown in Table. \ref{tab_98}.  

It can be seen that, compared with `Baseline',  `Baseline+Modality Prompts' significantly improves the performance. This may be due to the fact that the modality prompts can adaptively adjust the feature space of the feature extractor according to the characteristics of the input modalities, thus extracting discriminative unimodal features.  Then, by virtue of our proposed MTC loss, `Baseline+MAFE' can further learn more modality-distinguishable modality prompts in training, thus capturing more discriminative unimodal features from different modalities in  testing. Moreover, although significantly boosting performance, our proposed MAFE introduces minimal additional parameters and necessitates no modifications to its network structure as the number of modalities increases, thus maintaining efficiency and flexibility.
Finally,  `Baseline+MAFE+CSFH' achieves the best performance since the proposed CSFH effectively exploits the channel-wise and spatial-wise relations among multimodal features at different levels, thus effectively exploiting complementary information across modalities and further boosting performance.

\begin{table*}[!t]
	\renewcommand{\arraystretch}{1.3}
	\caption{Quantitative results of our proposed MSN with different types of inputs} 
	\label{tab_100}
	\centering
	\begin{tabular}{cccccccc}
		\hline 
		& RGB& D & T & RGB-D &RGB-T& D-T& RGB-D-T  \\
		\hline 
		$\mathcal{M}\downarrow$  &0.0052   &0.022   &0.0054  & 0.0043  &0.0041  &0.0051  &0.0035    \\  
		$F_\beta\uparrow$ 	     & 0.777   & 0.452  & 0.766  & 0.797   &0.817   &0.778  &0.845    \\  
		FLOPS(G)	& 28.1 &  28.1  &  28.1 & 38.3  &38.3 &38.3  &48.2    \\  
		FPS 	& 26.4& 26.4 & 26.4 & 17.4  &17.4 &17.4  &12.5   \\  
		\hline
	\end{tabular}
\end{table*}

\subsubsection{Quantitative Comparisons of CSFH under different SDFM and CDFM settings}  

In this subsection, we further evaluate our proposed CSFH by conducting several variants with different  SDFM and CDFM settings. Specifically, as depicted in Table \ref{tab_998}, we initially align the proposed SDFM with all levels of unimodal features. Subsequently, we gradually replace the SDFM with the proposed CDFM, commencing from the higher levels and progressing towards the lower levels.

It can be seen that compared with only employing SDFM or CDFM, aligning SDFM and CDFM to the features of different levels achieves better results. This validates the fact that the features at different levels have different characteristics. Low-level features have larger spatial sizes, thus using SDFM may extract more detail information by exploring spatial-wise interaction across modalities. While, high-level features have smaller spatial sizes with more semantics, thus employing CDFM may obtain better results by exploiting their channel-wise interactions. Furthermore, employing SDFM in the first two levels and CDFM in the last two levels achieves the best comprehensive performance. Therefore, we adopt this setting as our final model.

\subsubsection{Efficiency Comparisons}

In this subsection, we test our proposed MAT in sole mode and RGB-D-T setting of AM-XD dataset. Specifically, we evaluate our proposed MAT from RGB inputs, D inputs, T inputs, RGB-D inputs, RGB-T inputs, D-T inputs, and RGB-D-T inputs, respectively. Their results are shown in Table. \ref{tab_100}.

It can be seen that using more modalities usually obtains better saliency results than using fewer ones in most cases since more modalities will provide more scene information for identifying salient objects. However, it also indicates that using more modalities will significantly increase computational costs and reduce inference times. Specifically, using two modalities will introduce 36.2\% computational costs and reduce about 34.6\%  inference speeds than only employing one modality. And, employing three modalities will further increase about 25.8\% computational costs and reduce about 28.1\%  inference speeds than using two modalities. While the performance differences between one-modality inputs and two-modality inputs or between two-modality inputs and three-modality inputs may be relatively small for some particular combinations. Therefore, considering the balance between performance and efficiency, changing input type according to the conditions is necessary for most real-life applications, thus proving the importance of AM SOD tasks.


\section{Conclusion}

This paper proposes a novel modality-adaptive Transformer (MAT) for AM SOD. In the feature extraction stage of MAT, the proposed modality-adaptive feature extractor (MAFE) employs different modality prompts to adaptively tune the AM SOD model's feature space according to the characteristics of the input modalities. Meanwhile, the proposed modality translation contractive (MTC) loss facilitates MAFE learning more modality-distinguishable modality prompts during  training by pulling the images with the same modality prompts closer to each other and pushing the images with different modality prompts away from each other. By doing so, MAFE can effectively extract  discriminative unimodal features from different modalities, and tackle the diverse modality discrepancies. Furthermore, in the multimodal information fusion stage, the designed channel-wise and spatial-wise fusion hybrid (CSFH) strategy can effectively exploit cross-modal complementary information by exploring channel-wise or spatial-wise interactions of the unimodal features across modalities and levels. Eventually, our proposed MAT significantly boosts the performance of the AM SOD task.

\section*{Acknowledgments}

This work is supported by the China Postdoctoral Science Foundation under Grant No.2023M742745 and the Postdoctoral Fellowship Program of CPSF under Grant No.GZB20230559.
It is also supported by the National Natural Science Foundation of China under Grant No.61773301, the Shaanxi Innovation Team Project under Grant No.2018TD-012 and the State Key Laboratory of Reliability and Intelligence of Electrical Equipment No.EERI KF2022005, Hebei University of Technology.


\bibliographystyle{IEEEtran}
\bibliography{IEEEabrv,tex}
\begin{IEEEbiography}[{\includegraphics[width=1in,height=1.25in,clip,keepaspectratio]{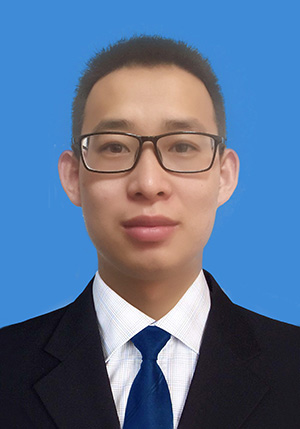}}]{Nianchang Huang}
	received the B. S. degree and the M. S. degree from Qingdao University of Science and Technology, Qingdao, China, in 2015 and 2018, and the Ph.D. degree in School of Mechano-Electronic Engineering, Xidian University, China, in 2022. He is currently a lecturer with the Automatic Control Department, Xidian University, China. His research interests include deep learning and multi-modal image processing in computer vision.
\end{IEEEbiography}
\vspace{-5mm}
\begin{IEEEbiography}[{\includegraphics[width=1in,height=1.25in,clip,keepaspectratio]{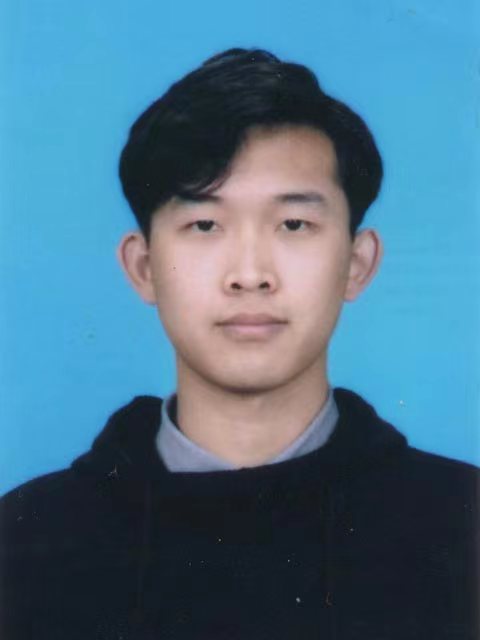}}]{Yang Yang}
	received his B. S. degree from Chang’an University, Xi’an, China, in 2019. He is currently pursuing the Ph.D. degree in School of Mechano- Electronic Engineering, Xidian University, China. His current research interests include multi-modal image processing and deep learning.
\end{IEEEbiography}
\vspace{-5mm}
\begin{IEEEbiography}[{\includegraphics[width=1in,height=1.25in,clip,keepaspectratio]{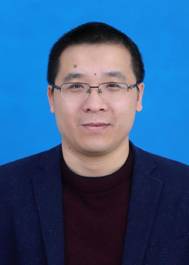}}]{Qiang Zhang}
	received the B.S. degree in automatic control, the M.S. degree in pattern recognition and intelligent systems, and the Ph.D. degree in circuit and system from Xidian University, China, in 2001,2004, and 2008, respectively. He was a Visiting Scholar with the Center for Intelligent Machines, McGill University, Canada. His current research interests include image processing, pattern recognition.
\end{IEEEbiography}
\vspace{-5mm}
\begin{IEEEbiography}[{\includegraphics[width=1in,height=1.25in,clip,keepaspectratio]{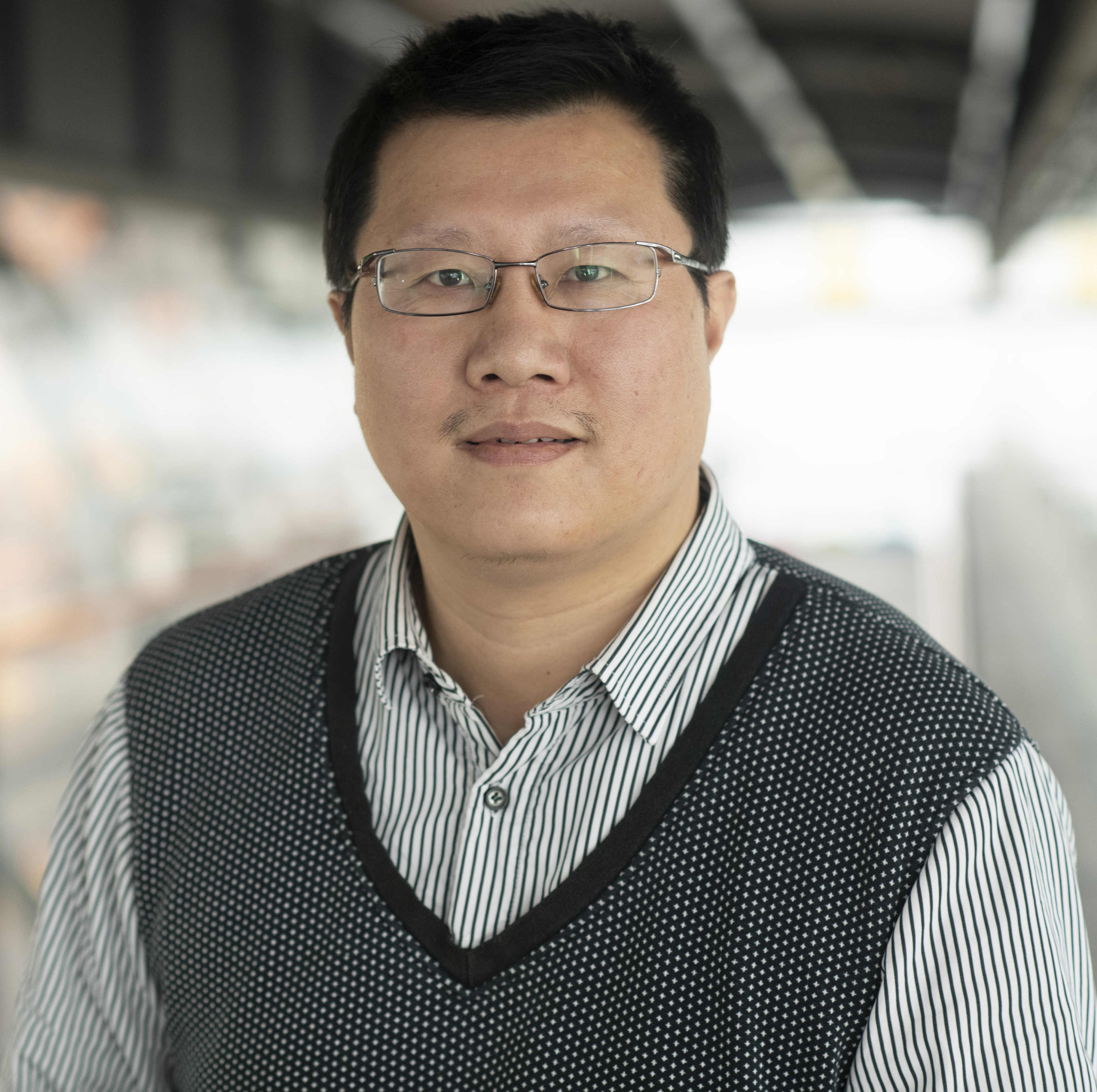}}]{Jungong Han}
	is Chair Professor in Computer Vision at the Department of Computer Science, the University of Sheffield, UK. He also holds an Honorary Professorship at the University of Warwick, UK. Previously, he was Chair Professor and Director of Research of the Computer Science department with Aberystwyth University, UK; Data Science Associate Professor with the University of Warwick; and Senior Lecturer in Computer Science with Lancaster University, UK. His research interests span the fields of video analysis, computer vision, and applied machine learning.
\end{IEEEbiography}
\begin{IEEEbiography}[{\includegraphics[width=1in,height=1.25in,clip,keepaspectratio]{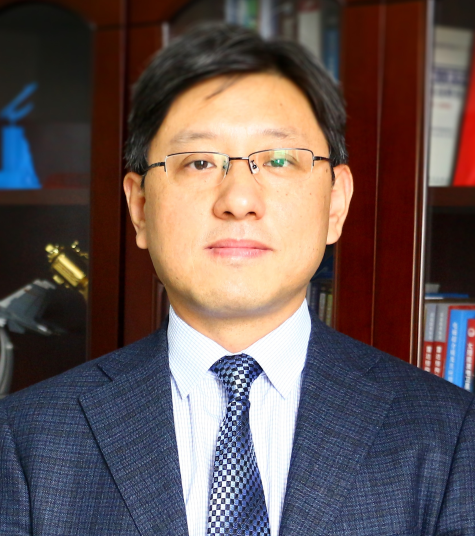}}]{Jin Huang}
	received the Ph.D. degree in mechanical engineering from Xidian University, Xi'an, China, in 1999. He worked with the Department of Mechanical Engineering, University of British Columbia, Vancouver, BC, Canada, as a Visiting Researcher in 2001-2002. He is a Professor, the Dean of the School of Electro-Mechanical Engineering, Xidian University, and the Director with the Key Laboratory of Electronic Equipment Design, Minister of Education. He also is the Fellow of the Chinese Institute of Electronics and served as the Deputy Secretary General of the Electro-mechanical Engineer Society of China. He has authored/co-authored more than 100 papers in various peer reviewed journals and conference proceedings, and holds more than 50 patents. His research interests include  exible electronics, mechatronics, and 3-D printing.
\end{IEEEbiography}
\end{document}